\documentclass[final]{cvpr}

\usepackage{times}
\usepackage{epsfig}
\usepackage{graphicx}
\usepackage{amsmath}
\usepackage{amssymb}

\usepackage{cite}
\usepackage{amsfonts}
\usepackage{algorithmic}
\usepackage{graphicx}
\usepackage{textcomp}
\usepackage{xcolor}
\usepackage[ruled,vlined]{algorithm2e}
\usepackage{float}
\usepackage{tabularx,ragged2e}
\usepackage{booktabs}
\newcolumntype{C}{>{\Centering\arraybackslash}X} 
\usepackage{dblfloatfix}
\usepackage{gensymb}

\newlength{\tempheight}
\newlength{\tempwidth}

\newcommand{\rowname}[1]
{\rotatebox{90}{\makebox[\tempheight][c]{\textbf{#1}}}}

\newcommand{\columnname}[1]
{\makebox[\tempwidth][c]{\textbf{#1}}}

\usepackage[pagebackref=true,breaklinks=true,colorlinks,bookmarks=false]{hyperref}

\def\cvprPaperID{3} 
\def\confYear{CVPR 2021}

\usepackage{comment}
\usepackage{subfig}
\usepackage{multirow}
\usepackage{adjustbox}
\usepackage{nopageno} 

\begin{document}

\title{GOO: A Dataset for Gaze Object Prediction in Retail Environments}

\author{Henri Tomas$^{1}$\thanks{Equal contribution.}\qquad Marcus Reyes$^{1}$\footnotemark[1]\qquad Raimarc Dionido$^{1}$\footnotemark[1]\qquad Mark Ty$^{1}$\\ Jonric Mirando$^{1}$\qquad Joel Casimiro$^{1}$\qquad Rowel Atienza$^{1}$\qquad Richard Guinto$^{2}$\\
\\
$^{1}$University of the Philippines \qquad $^{2}$Samsung R\&D Institute Philippines\\
{\tt\small {\{henri.tomas,	marcus.joseph.reyes, raimarc.dionido, mark.vincent.ty,}}\\
{\tt\small{jonric.mirando, 	joel.casimiro, 	rowel\}}@eee.upd.edu.ph},  {\tt\small{rfguinto@samsung.com}}
}

\maketitle

\begin{abstract}

One of the most fundamental and information-laden actions humans do is to look at objects. However, a survey of current works reveals that existing gaze-related datasets annotate only the pixel being looked at, and not the boundaries of a specific object of interest. This lack of object annotation presents an opportunity for further advancing gaze estimation research. To this end, we present a challenging new task called gaze object prediction, where the goal is to predict a bounding box for a person's gazed-at object. To train and evaluate gaze networks on this task, we present the Gaze On Objects (GOO) dataset. GOO is composed of a large set of synthetic images (GOO-Synth) supplemented by a smaller subset of real images (GOO-Real) of people looking at objects in a retail environment. Our work establishes extensive baselines on GOO by re-implementing and evaluating selected state-of-the-art models on the task of gaze following and domain adaptation. Code is available\footnote[1]{\url{https://github.com/upeee/GOO-GAZE2021}} on github.


\end{abstract}

\section{Introduction}
Everywhere we go, we see people looking at objects. Knowing what someone is looking at often gives information about that person. Someone looking at a map might be a tourist looking for directions. A person looking at the traffic light is probably planning to cross the street. In retail, a salesperson who can identify the product a customer is looking at can quickly offer assistance. Where and what we look at potentially reveals something about us and what we're doing.

Emery\cite{emery_eyes_2000} showed the neuro-scientific importance of gaze by elaborating on how it is used for social interaction, for indicating intention, and for communication between people. Similarly, gaze can also be a crucial factor for computer vision systems in understanding and interpreting human actions in a certain scenario. Recasens \etal \cite{recasens_where_2015} defined the task of gaze following for these systems as that of determining the direction and the point a person is looking at. 

The potential applications of intelligent systems with the ability to do gaze following lead to increased interest in varying gaze-related subfields. Several datasets are created for predicting saliency \cite{borji_cat2000_2015, jiang_salicon_2015, xu_turkergaze_2015}, or determining portions of an image that is most likely to catch interest from a first person point-of-view. Another subfield exists for tracking eye-movement to predict the gaze direction from a second person perspective \cite{funes_mora_eyediap_2014}. Gaze prediction on humans in images viewed from third-person became the most commonly researched subfield, after well-established baselines were published using the GazeFollow dataset \cite{recasens_where_2015}. Subsequent works \cite{lian_believe_2019, chong_detecting_2020} applied deep neural networks to achieve near-human performance on this task, and developed methods that can track human gaze in video. 

\begin{figure*}[t!]
    \centering 
    \setlength{\tempwidth}{0.30\textwidth}
    \settoheight{\tempheight}{\includegraphics[width = 0.25\textwidth]{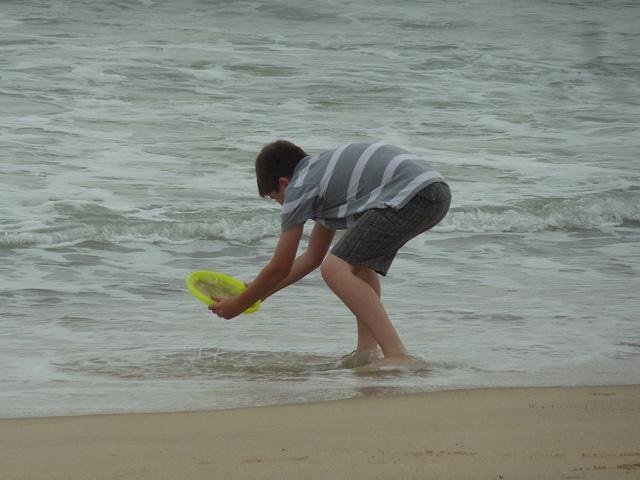}}
    \rowname{GazeFollow}
    \subfloat{\includegraphics[width = 0.30\textwidth, height = 0.15 \textheight]{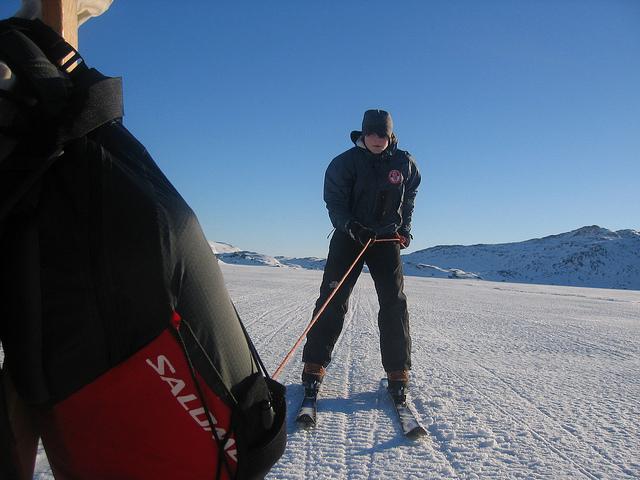}}
    \subfloat{\includegraphics[width = 0.30\textwidth, height = 0.15\textheight, ]{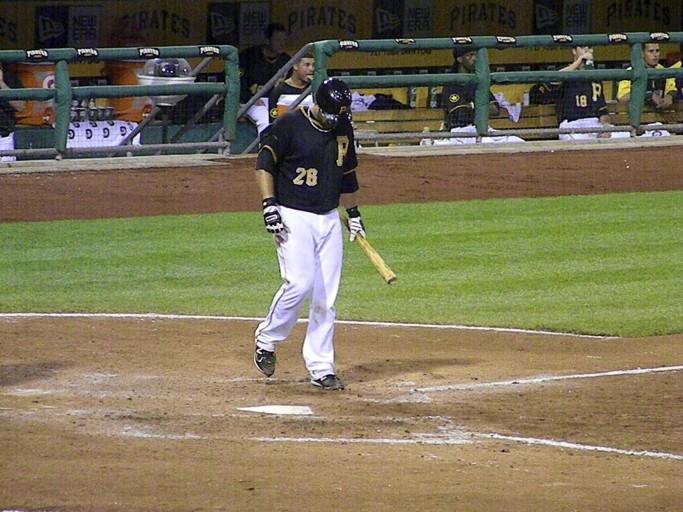}}
    \subfloat{\includegraphics[width = 0.30\textwidth, height = 0.15\textheight, ]{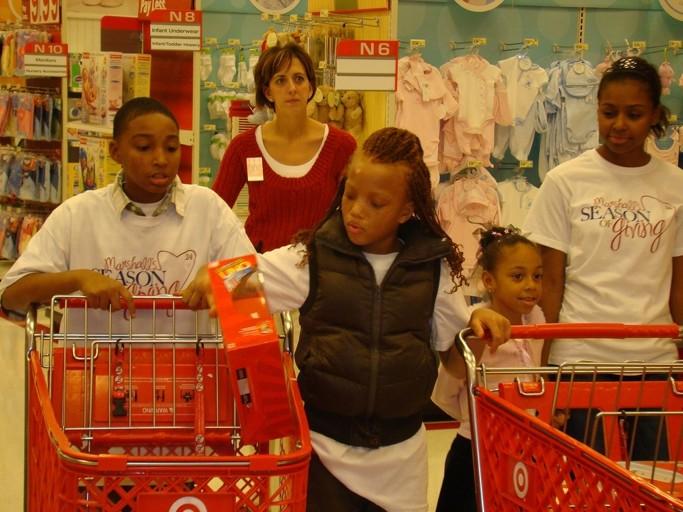}}\\[-1.75ex]
    \setlength{\tempwidth}{0.30\textwidth}
    \settoheight{\tempheight}{\includegraphics[width = 0.30\textwidth]{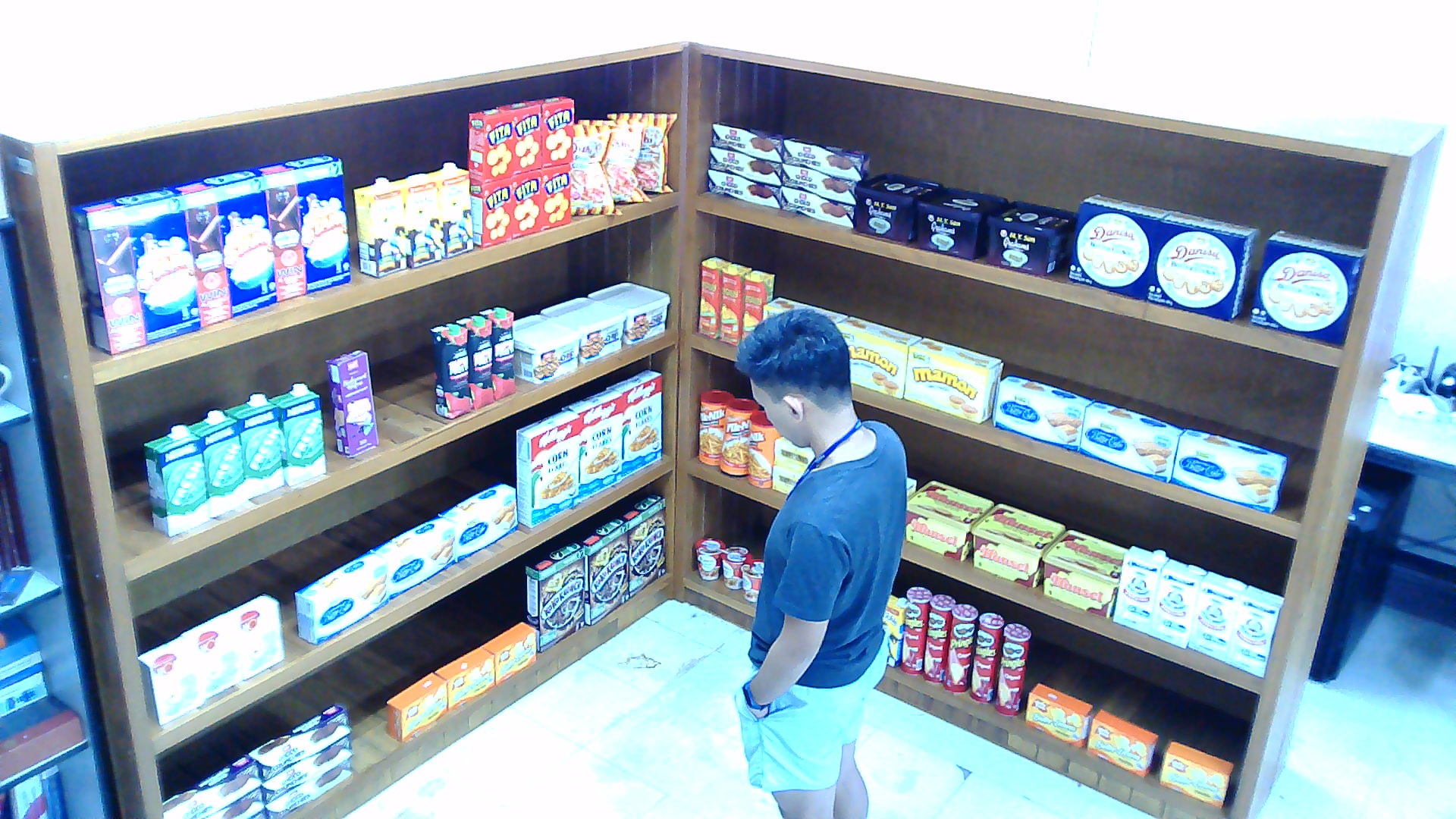}}
    \rowname{GOO-Real}
    \subfloat{\includegraphics[width = 0.30\textwidth, height = 0.15\textheight, ]{images/Real_Synth_GazeFollow/real1.jpg}}
    \subfloat{\includegraphics[width = 0.30\textwidth, height = 0.15\textheight, ]{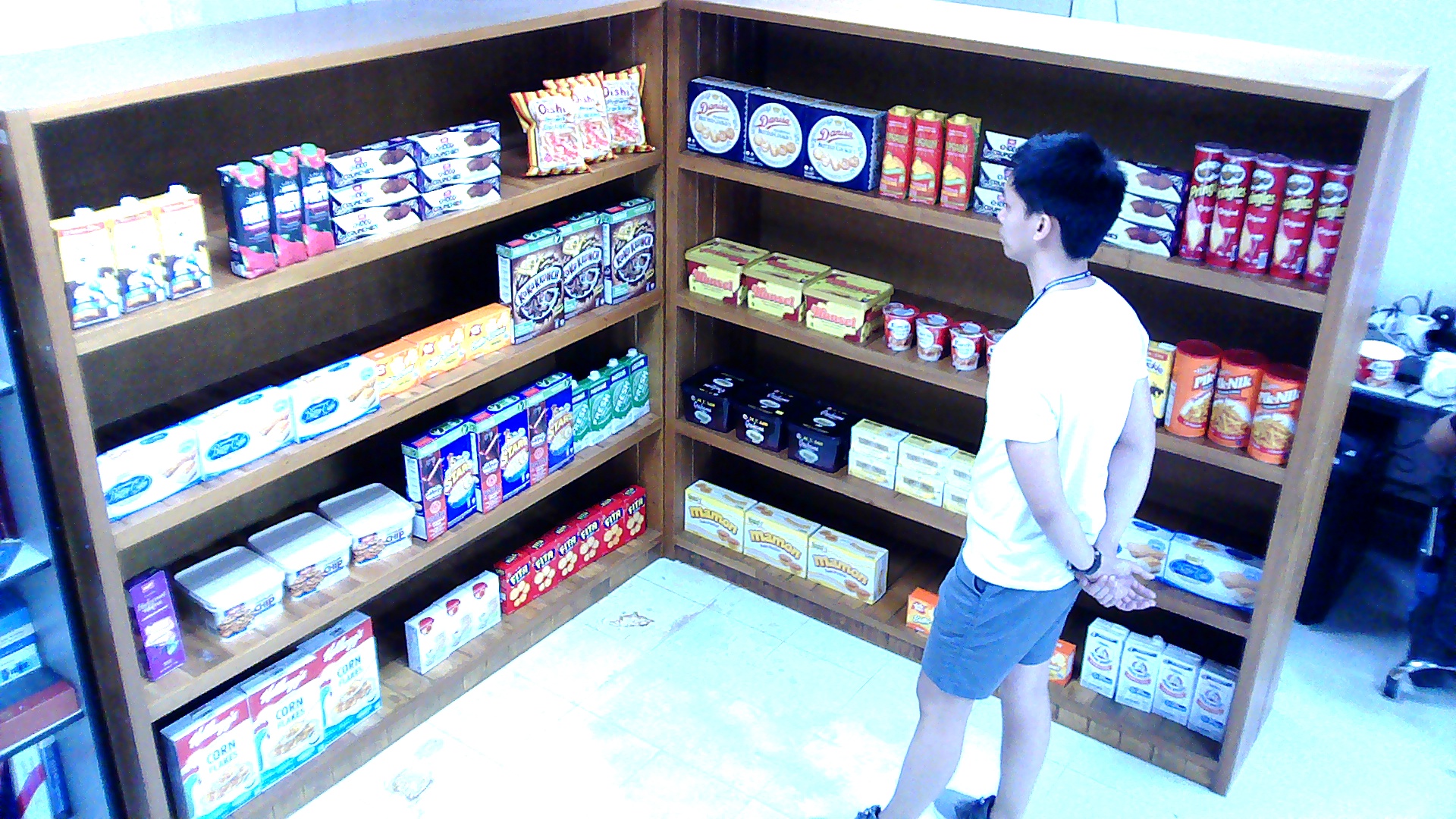}}
    \subfloat{\includegraphics[width = 0.30\textwidth, height = 0.15\textheight, ]{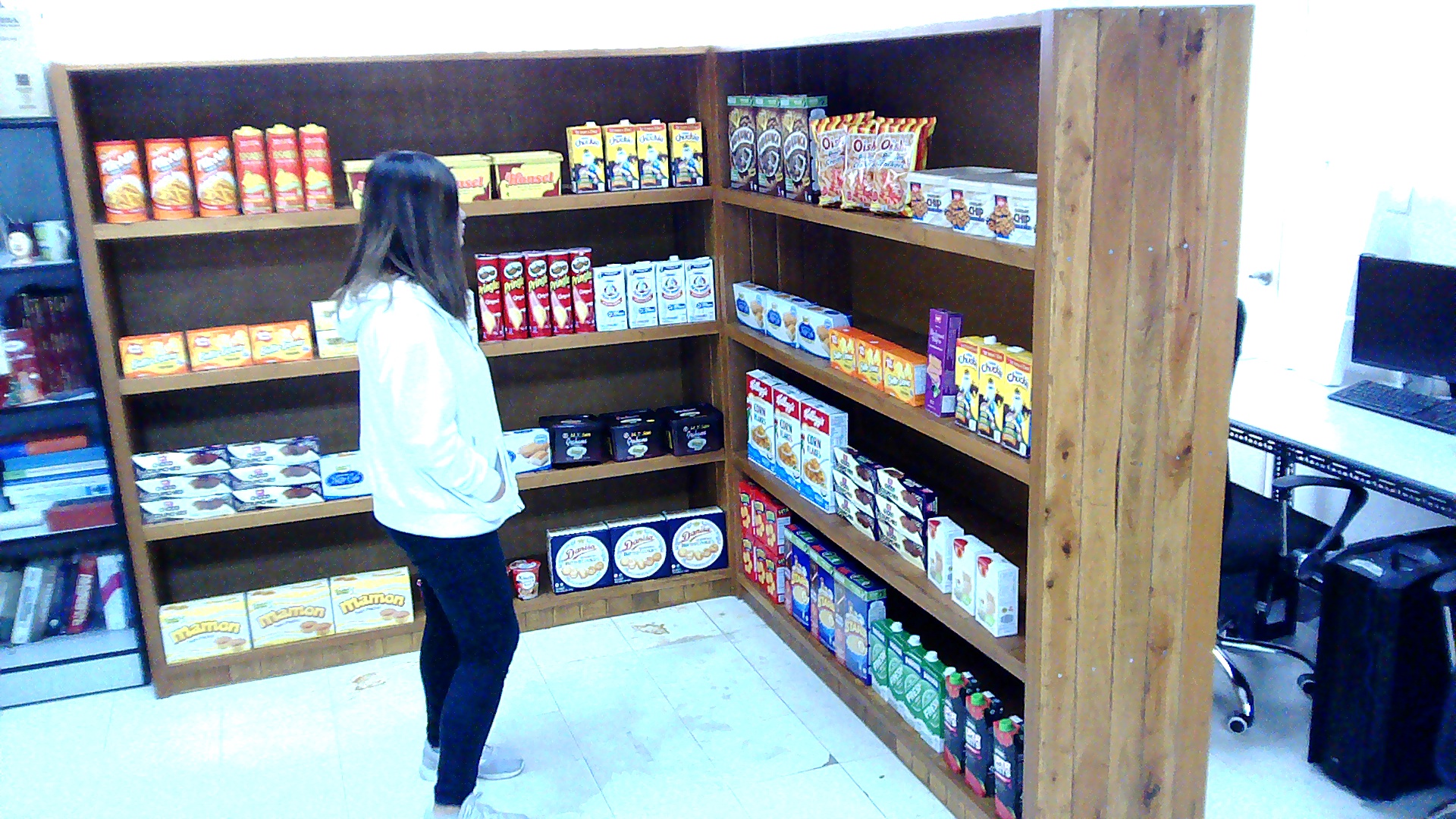}}\\[-2.35ex]
    \setlength{\tempwidth}{0.30\textwidth}
    \settoheight{\tempheight}{\includegraphics[width = 0.30\textwidth]{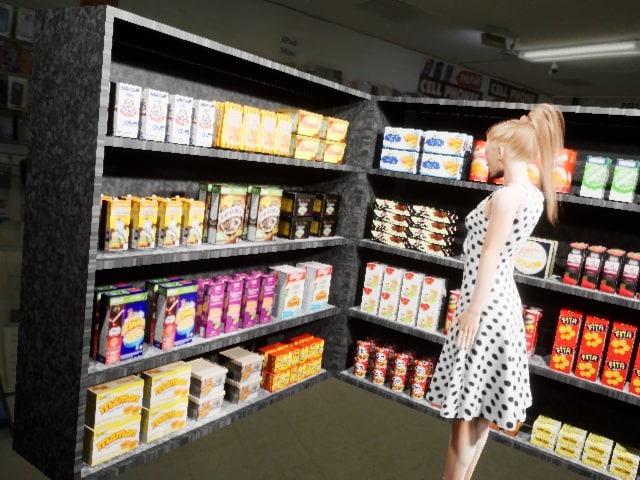}}
    \rowname{GOO-Synth}
    \subfloat{\includegraphics[width = 0.30\textwidth, height = 0.15\textheight, ]{images/Real_Synth_GazeFollow/synth1.jpg}}
    \subfloat{\includegraphics[width = 0.30\textwidth, height = 0.15\textheight, ]{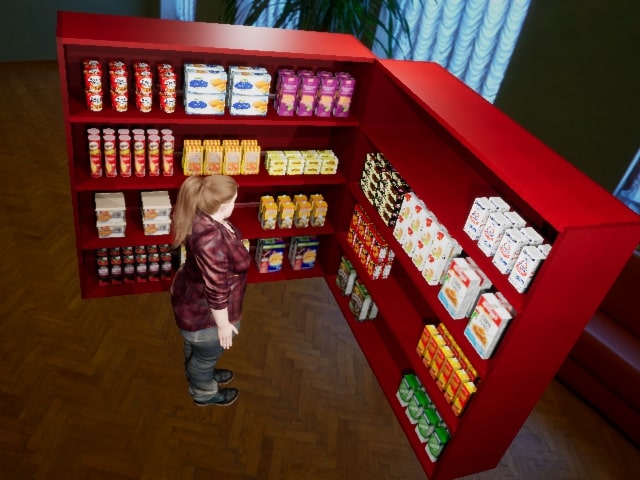}}
    \subfloat{\includegraphics[width = 0.30\textwidth, height = 0.15\textheight, ]{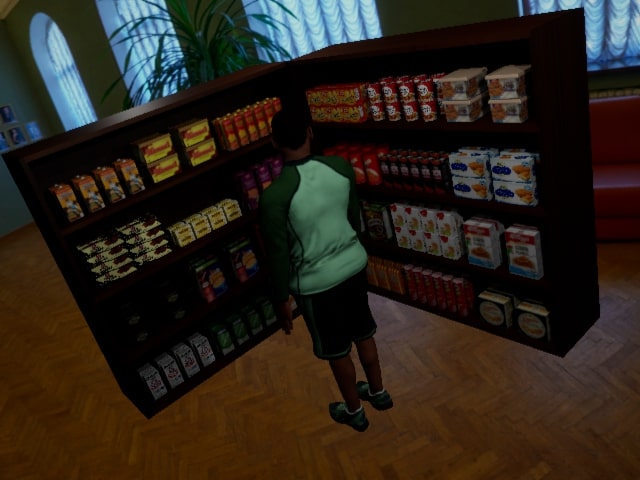}}\\
    \caption{Samples of images from GazeFollow (1st row), GOO-Real (2nd row), and GOO-Synth (3rd row).}
    \label{fig:sample1}
\end{figure*}

Taking inspiration from how humans perform gaze following, we believe that identifying which object a person is looking at holds more value than predicting a point. When you follow another person's gaze, it seems natural to take into account the objects in the inferred direction to confirm where and what exactly this person is looking at. Similarly, teaching a system to be aware of the objects in a scene could aid gaze following, and may result in more accurate predictions.

To this end, we present a new task called \emph{gaze object prediction}, where one must infer the bounding box of the object gazed at by the target person, which will be referred to as the \emph{gaze object}. Aside from being more challenging, the task also encourages the use of objects present in the scene to build better performing gaze systems. In environments with fewer objects, sparse object placement can be used as cues for the model to affirm whether the estimated direction is correct. Conversely, in environments with dense object placement, clustering of objects may hold important features that the model can learn to be more robust in its predictions. Our work focuses on the prediction of gaze objects in retail, a task of fulfilling both sparse and dense conditions, with promising applications in market research. We demonstrate that existing gaze-related datasets lack the annotation required for training on our proposed task.


To address this problem, we present a new image dataset called Gaze On Objects (GOO), a dataset tailored for gaze object prediction in retail environments. It is composed of synthetic and real images, and is considerably larger than existing datasets. Aside from the standard gaze annotation such as gaze point and the person's head, GOO includes additional detailed annotations such as bounding boxes, classes, and segmentation masks for every object in the image. Its differences with  GazeFollow, which is a favored dataset for evaluation on predicting gaze points in third person, is discussed in detail at section \ref{comparingGazeFollow}.

We also establish comprehensive baselines on the GOO dataset by evaluating existing state-of-the-art gaze networks on the task of gaze following. 
Lastly, to provide insight into how GOO can be used for domain adaptation, experiments on the transferability of GOO's synthetic features to the real domain is provided. 

\section{Related Work}

In the following, we discuss related datasets and justify why they are not suitable for the task of predicting gazed-upon objects.

iSUN\cite{xu_turkergaze_2015}, a subset of SUN\cite{xiao_sun_2010} is annotated with first person saliency heatmaps based on eye tracking. It is a small dataset composed of 20,608 images.

CAT2000 is a compilation of various datasets (one if which is also SUN\cite{lin_microsoft_2015}). With only 4,000 images, it is much smaller than iSUN\cite{xu_turkergaze_2015}. It is annotated with first person saliency heatmaps via eye tracking. 

SALICON\cite{jiang_salicon_2015}, a subset of MSCOCO\cite{lin_microsoft_2015} composed of 10,000 images, is annotated with first person saliency heatmaps via mouse tracking. 

EYEDIAP\cite{funes_mora_eyediap_2014} is a video dataset in a second person view of a person's face. This person is the one whose gaze is being predicted. There are 16 different participants with ~4 hours of data. The gazed upon object is either floating in front of the camera (visible in the video frame) or on a screen behind the camera (not visible in the video frame).

None of these are suited for the task of predicting which object a human is looking at given only a third person view. This is primarily due to the different perspectives these datasets were captured in. Furthermore, some of them do not even have the ground truth gaze annotations as Gorji \etal \cite{gorji_attentional_2017} had to manually add these for their work on augmented saliency heatmaps. Finally, it is also worth noting that these datasets are all very small with the largest image dataset containing only around 20,000 photos.

\begin{table}[t] \centering
\begin{tabular}{c|ccc}
\hline
\textbf{Dataset} & \textbf{\begin{tabular}[c]{@{}c@{}}Ground\\ Truth\end{tabular}} & \textbf{Perspective} & \textbf{Size} \\ \hline
iSUN \cite{xiao_sun_2010}             & Point                                                              & 1st Person           & 20,000        \\
SALICON \cite{jiang_salicon_2015}         & Point                                                              & 1st Person           & 10,000        \\
CAT2000 \cite{borji_cat2000_2015}          & Point                                                              & 1st Person           & 4,000         \\
EYEDIAP \cite{funes_mora_eyediap_2014}         & Point                                                              & 2nd Person           & N/A           \\
GazeFollow \cite{recasens_where_2015}      & Point                                                              & 3rd Person           & 122,143       \\
GOO (Ours)       & Object                                                             & 3rd Person           & 201,552       \\ \hline
\end{tabular}
\caption{Survey of saliency and gaze-related datasets. Previous datasets are small in terms of size save for GazeFollow, and only GOO (ours) has annotations for the gaze object bounding boxes.} \label{fig:all-datasets.}
\vspace{-8px}
\end{table}

GazeFollow\cite{recasens_where_2015} is currently the most suitable dataset for the gaze following subfield which we focused on. This dataset was published by Recasens \etal along with a gaze heatmap prediction system, and has been used by other gaze prediction methods such as that of Chong \etal \cite{chong_detecting_2020} and Lian \etal \cite{lian_believe_2019}. It is composed of 122,143 images compiled from various preexisting datasets, which were then annotated with ground truth gaze point locations. Thus, it is built for gaze point prediction, and not for gaze object prediction. 

The task for which they were designed is the main differentiator between GazeFollow and GOO. This alone is not enough to warrant the creation of a new dataset; after all, it can be argued that GazeFollow could just be annotated with ground truth gaze objects. Therefore, we enumerate more differences between Gazefollow and GOO in their annotations, size, context, suitability to task, and domain adaptation applications (See summary in Table \ref{tab:table_comp}). We further explain the differences in Section \ref{comparingGazeFollow}.


In our work, we employ existing gaze following methods \cite{chong_detecting_2020, lian_believe_2019, recasens_where_2015}  with established baselines on the GazeFollow dataset. In Section \ref{baselinemethods}, we will discuss more thoroughly what these works are and how we recreated and benchmarked them on both GazeFollow and GOO. Finally, we mention that to the extent of our research there is no work yet that is specific for the task of gaze object prediction.

\begin{table}
    \centering
    \begin{tabular}{c|cc}
    \hline
    & \textbf{GazeFollow} & \textbf{GOO}\\
    \hline
    \hline
    \textbf{Size} & 122,143 & 201,552\\
    \hline
    \textbf{Type} & Real & Synthetic \& Real\\
    \hline
    & Head Bbox, & Head Bbox,\\
    \textbf{Annotations} & Gaze point & Gaze object,\\
    & & Obj Segmentation\\
    \hline
    \textbf{Context} & Varied & Retail\\
    \hline
    \textbf{Ppl./image} & Varied & 1\\
    \hline
    \textbf{Obj./image} & Few & Many \\
    \hline
    \textbf{Applicable for DA} & & \checkmark \\
    \hline
    \end{tabular}
    \caption{Differences between GazeFollow and GOO.}
    \label{tab:table_comp}
\vspace{-8px}
\end{table}

\begin{figure*}[h]
\begin{centering}
\includegraphics[width=\linewidth]{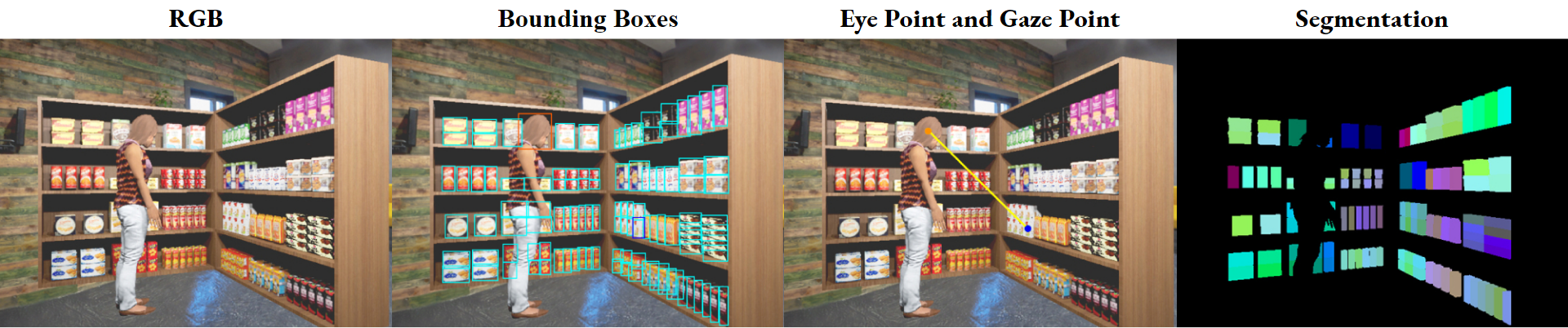}
\end{centering}
\vspace{-12px}
\caption{\protect Annotations for the GOO dataset. From left to right: RGB image, bounding boxes with object class, eye point and gaze point, and segmentation masks. Bounding boxes and segmentation masks for the head and gaze object are indicated. \label{fig:annotations}}
\end{figure*}

\section{Gaze On Objects (GOO)}

The GOO dataset is composed of images of shelves packed with 24 different classes of grocery items, where each image contains a human or a human mesh model gazing upon an object. All objects in the scene are annotated with their bounding box, class, and segmentation mask. As with existing gaze-related datasets, location and bounding box annotations for the person's head are provided. With these annotations, GOO can also be used for other tasks such as object detection and segmentation.

GOO Dataset consists of two parts: a larger synthetic set of images called GOO-Synth, and a smaller real set of images called GOO-Real. 

\subsection{GOO-Real}
A mock-up of a retail environment was built. Several grocery items were placed on the shelves to imitate a real grocery store. GOO-Real consists of 100 humans (68 male and 32 female ranging from 16 to 50 years old) and 9,552 images. For each image there are around 80 total grocery items, each belonging to one of 24 different classes. The shelves are completely filled up by 3 to 6 instances of the same grocery item. Two cameras were used, one facing each cabinet. For each volunteer the items were shuffled to avoid overfitting when training models. The test set is made up of 2,156 images with the remaining images comprising the training set, split so that human volunteers in one do not appear in the other.

For the creation of GOO-Real, videos were taken of each volunteer. Each volunteer was asked to walk into the simulated grocery environment. They would then be told to gaze at a total of 24 items for a few seconds each. Two images were extracted from the video for each item stared at. A predetermined randomized list was used to instruct each volunteer regarding which specific item he should look at (\eg Look at the box of cereal located at shelf 1, row 2, 2nd from the left). These lists were later used by 11 annotators when attaching ground truth labels assuring that the objects being labelled as ground truth objects were indeed the items being gazed upon.

\subsection{GOO-Synth}
GOO-Synth forms the bulk of GOO's training data with 192,000 images, to which the smaller GOO-Real will be supplementary. For the creation of GOO-Synth, a realistic-looking replica of the scene used in GOO-Real was created in Unreal Engine \cite{unrealengine}. Five cameras (randomly chosen from 50 virtual cameras placed inside the simulated environment) was used to capture images of one of 20 synthetic human models interacting with the scene. These human models were highly varied with respect to skin tone (black, white, brown, etc.), gender (male, female), body form (fat, thin, muscular, tall, short, etc.), and outfits. The grocery objects were designed after real-life counterparts, with the packaging of real objects scanned to be used as textures. Other elements of the scene were also varied such as skyboxes (background), and lighting. In total, there were 38,400 scene environments. 

To simulate the act of looking, we created a gaze vector originating from a point between the eyes of the human model and perpendicular to the face. This gaze vector was directed towards the indicated ground truth object. Similar to GOO-Real, the human models were split such that each human model appeared exclusively in the training set or the test set. The training set used 18 models, while the test set used the remaining two models.

\subsection{Comparing to GazeFollow\label{comparingGazeFollow}}

\textbf{Annotations.} For GazeFollow \cite{recasens_where_2015}, the gaze point annotations were added manually; any additional annotations such as object bounding boxes will also have to be done manually. On the other hand, since the bulk of GOO is synthetic, annotation is not only easier but also faster since the task can be automated. Another difference is that annotations for GOO has better integrity. This is due to the ground truth object being noted down in advance before human volunteers or models are made to look at it, as opposed to GazeFollow which sets the gaze point ground truth based on the judgement of volunteer annotators.

\textbf{Size and Domain.} Our dataset is much larger than the GazeFollow dataset, with GOO having 201,552 images compared to GazeFollow's 122,143 images. This is due to the bulk of GOO being synthetic, where unique images can be generated by adjusting conditions in the simulated environment. It should be noted that only 9,552 samples of GOO are real images, compared to GazeFollow which is entirely real-world data; therefore, performance of models trained on GOO in real scenarios depend mostly on how well it can adapt learned synthetic features. We discuss more of GOO for domain adaptation in section \ref{domadapt}.


\textbf{Context.} Our dataset is focused on the retail environment. GazeFollow was built by the authors from a variety of other datasets which are not necessarily suited for one particular setting \cite{recasens_where_2015}. In contrast, GOO is tailor-made for the task of object gaze prediction in a densely-packed environment. While we do not claim that retail is the only environment that would benefit from gaze following, we believe that it is one of the fields where the advantages are very apparent. For example, most grocery stores already have the equipment in the form of security cameras. Furthermore, in a retail setting, knowing what objects hold interest is useful.

\textbf{Suitability to Task.}  GazeFollow consists of images borrowed from a combination of different datasets. In these datasets, there is a prevalence of scenes where objects are few and sparsely placed. GOO's retail setting provides an aspect which GazeFollow generally does not, and that is gaze estimation in an image densely packed with objects. The task of predicting which object is being gazed at in scenes with many objects is inherently harder when compared to scenes with fewer objects. However, we hypothesize that models trained with dense objects are more likely to learn important features making it more robust in its predictions.





\subsection{Tasks} \label{tasks}

The extensive annotation of the GOO dataset makes it applicable to training systems on a multitude of challenging problems, especially along the fields of gaze estimation and object detection. In this paper we highlight the applications of GOO on three tasks, which we define as follows.

\textbf{Gaze following.} The task of gaze following as defined by Recasens \etal \cite{recasens_where_2015} entails the prediction of the exact point a person is looking at, given the image and the head location. The task can be broken down into two stages, namely: 1) the estimation of gaze direction from the head and scene features and 2) the regression of confidence values for a gaze point heatmap. The GOO dataset can provide benchmarks on this task by defining the ground truth object's center as the gaze point. 

\textbf{Gaze Object Prediction.} The action of predicting the gaze point remains a challenging problem. However, in practical applications such as identifying the object being looked at, current works trained on estimating a single point would require separate systems for classification and detection. We propose a novel task called gaze object prediction: the goal is for an intelligent system to learn to classify and predict boundaries for the object a person is looking at. We believe this  presents a much more challenging problem compared to gaze following, as learning features that are important to gaze must be balanced with features tantamount to object detection. The GOO dataset's scope lies on applying this task to retail environments, where multiple products in close proximity provide difficult yet rewarding samples for a model to learn from. However, the current works on gaze estimation do not predict the gaze objects. Thus, we will leave performance measurements for gaze object prediction for future work.

\textbf{Domain Adaptation.}\label{domadapt} Considering that GOO is composed of a .95 to .05 split between synthetic images and real images, exploring how well features learned on GOO-Synth can adapt to the domain of GOO-Real is also a problem that merits interest. We benchmark the gaze prediction networks trained with the GOO-Synth dataset on the task of domain adaptation, specifically on transferring the learned features from the synthetic domain onto the real domain. This task evaluates the performance of the baselines when trained with simple transfer learning on the GOO-Real dataset, comparing architectures with prior training on GOO-Synth to those without.

\section{Methodology} \label{methodology}

In this section we discuss the methods selected to provide benchmarks on the GOO dataset, along with the criteria followed in choosing these methods. A comprehensive discussion of each baseline is provided, where stages and techniques are outlined to give insight into how the task of gaze following is accomplished in a modular fashion. 

\subsection{Baseline Selection}

To verify the accuracy of our implementation of the baselines, it is highly beneficial to have an existing performance benchmark on another dataset to serve as a point of comparison. The GazeFollow dataset is an important cornerstone of the gaze following task, and a considerable amount of state-of-the-art methods already have a benchmark on this dataset; thus, we use these benchmarks to guarantee the correctness of our implementation of the baselines before evaluating on the GOO dataset.

The input to the network architectures should only include the full input image along with the head location. This criteria rules out methods that use video, preceding frames, or 3D annotations as supporting data for the gaze prediction. However, such methods that can be modified to follow correct inputs can be considered. The output of the baselines should include a final gaze heatmap of no specific dimensions. The point in the heatmap with the highest confidence value shall indicate the gaze point, and both heatmap and gaze point are used for the evaluation on the previously discussed tasks.

\begin{figure}[t]
\begin{centering}
\includegraphics[width=1\columnwidth]{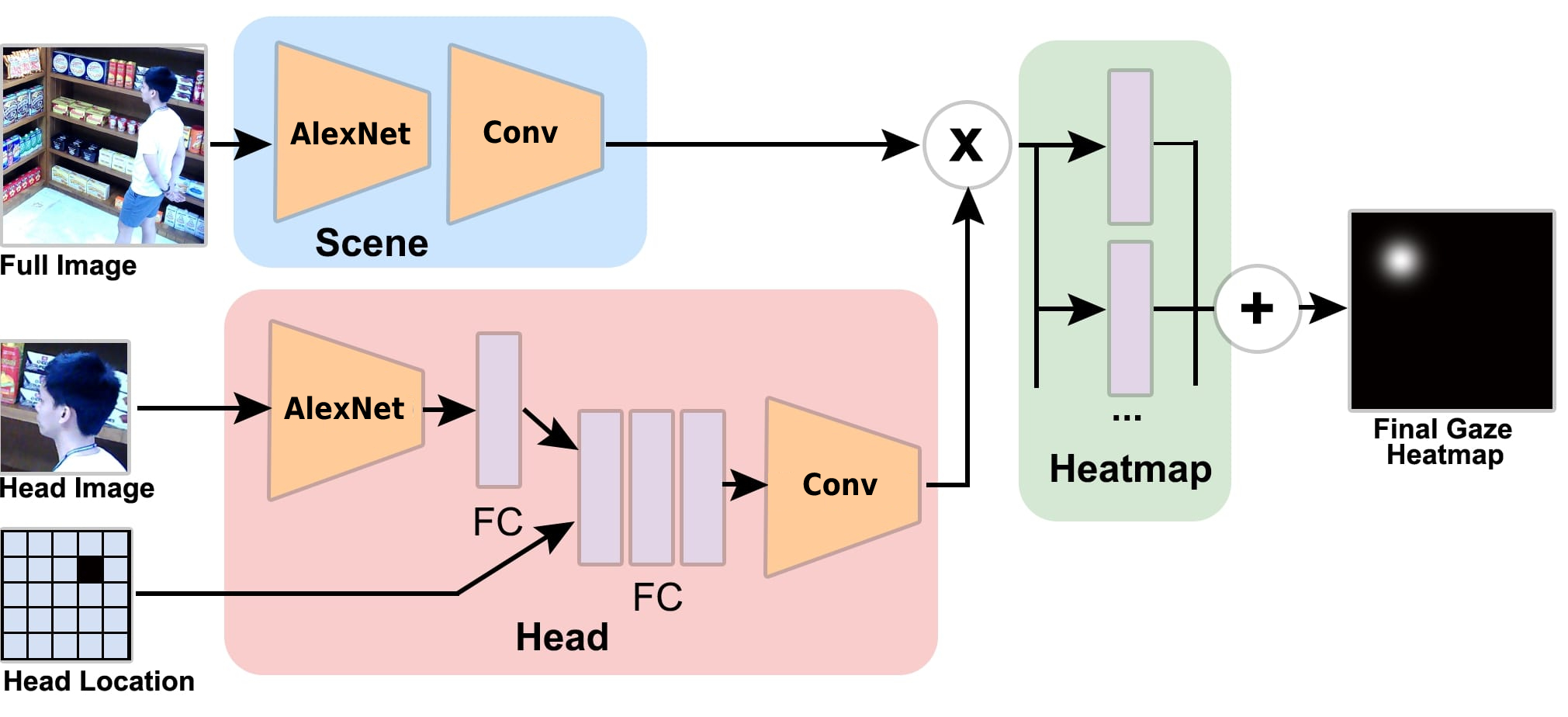}
\par\end{centering}
\vspace{-5px}
\caption{\protect Where are They Looking? by Recasens \etal \cite{recasens_where_2015} \label{fig:recasens_diag}}
\vspace{-10px}
\end{figure}

\subsection{Baseline Methods} \label{baselinemethods}

Considering the above criteria, the works of Recasens \etal \cite{recasens_where_2015}, of Lian \etal \cite{lian_believe_2019}, and of Chong \etal \cite{chong_detecting_2020} are selected as baselines to be evaluated on the GOO dataset. The contrived architectures of these works set the precedent of dividing the task of gaze following into three sub-problems, to be solved by different modules. We define both the modules and their respective sub-problems and enumerate them as: 1) the scene module, which performs feature extraction on the entire image; 2) the head module, which performs feature extraction on the cropped head image and location; and 3) the heatmap module, which uses the scene and head feature maps to predict a gaze point confidence heatmap. Each network architecture discussed in this section is visualized in terms of these three modules.

\textbf{Random.} When quantitatively benchmarking the performance of multiple networks, it would be best to have a lower bound for performance. For this we establish the same random baseline used by \cite{recasens_where_2015}, where a heatmap is generated per pixel by sampling values from the standard normal distribution. This heatmap is then treated as the output heatmap and evaluated against the ground truth.

\textbf{Where are they Looking?}. An architecture for gaze following can be observed in Figure \ref{fig:recasens_diag}, representing the work of Recasens \etal \cite{recasens_where_2015}. Their work sets a precedent in their approach of having two distinct input pathways: one module for the full image and another module for the cropped head image. They design the scene module inspired by saliency networks, which highlights important subjects in the image, including objects that a person might look at. The head module is then designed to infer the general direction of the person's gaze. Both of these modules use AlexNet \cite{krizhevsky_imagenet_2012} for feature extraction, which uses pretrained weights for ImageNet \cite{russakovsky_imagenet_2014} and the Places dataset \cite{zhou_learning_2015} to initialize the head and scene module respectively.

The feature maps from the first two modules are then combined using element-wise multiplication. The resulting product is passed onto the network's heatmap module, marked as the green module in Figure \ref{fig:recasens_diag}. To produce the final heatmap, their work uses a shifted-grids approach, dividing the full image into five N $\times$ N grids of different ratios where each cell is treated as a binary classification problem (if the cell contains the gaze point). Per shifted grid, a fully-connected layer predicts confidence values for each cell, and the outputs from predicting on multiple grids are merged to form the final gaze heatmap. 

\begin{figure}[t]
\begin{centering}
\includegraphics[width=0.6\columnwidth]{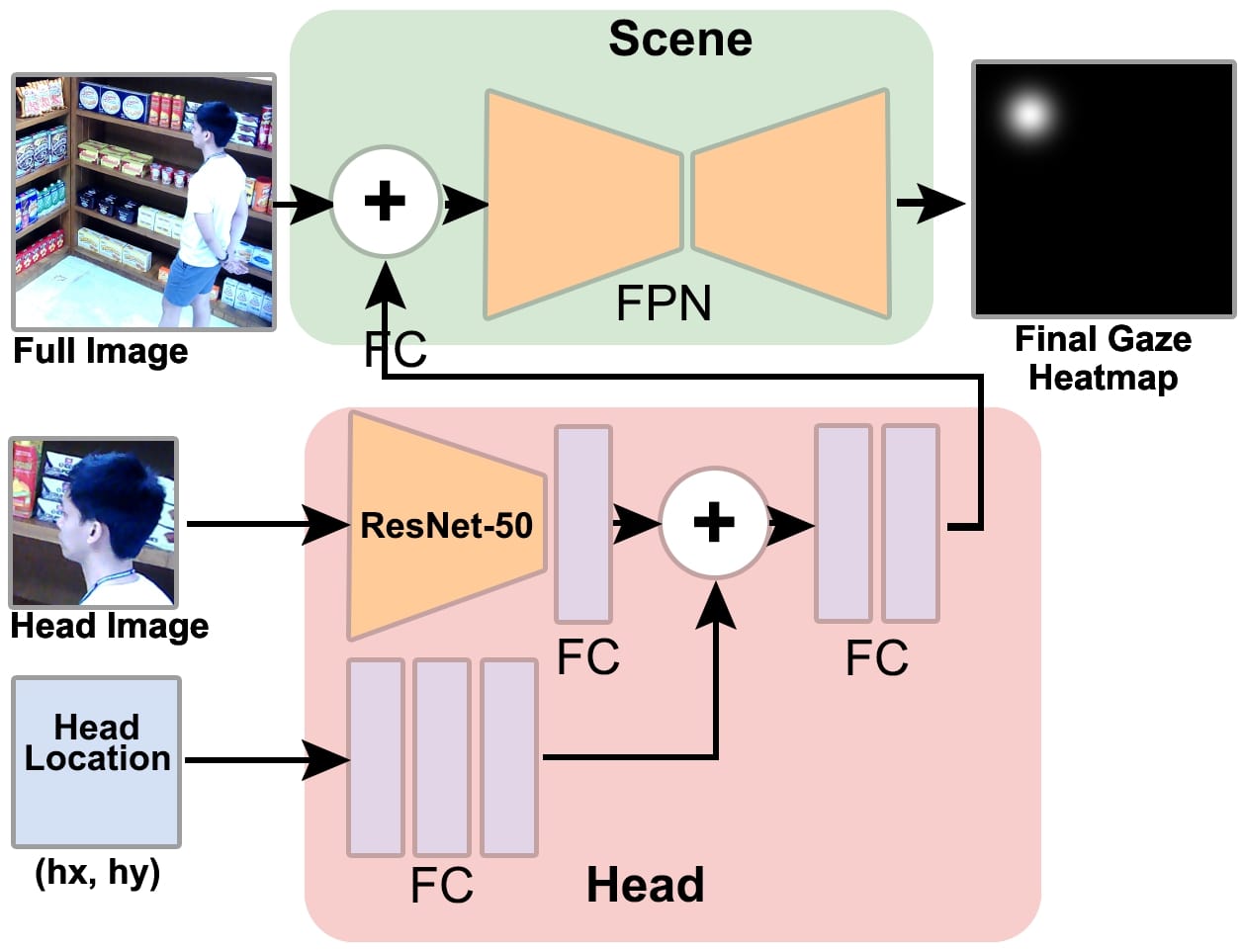}
\par\end{centering}
\vspace{-5px}
\caption{\protect Believe It or Not, We Know What You are Looking At! by Lian \etal \cite{lian_believe_2019} \label{fig:lian_diag}}
\vspace{-5px}
\end{figure}

\textbf{Believe It or Not, We Know What You are Looking At!}. Subsequent work conducted by Lian \etal \cite{lian_believe_2019} introduced state-of-the-art CNNs in gaze networks. They proposed a new architecture as seen in Figure \ref{fig:lian_diag} where the head module infers the gaze direction from the head image using ResNet-50 \cite{he_deep_2016}. The head location is encoded by fully-connected layers before being concatenated with the head feature map. Instead of producing a directional gaze mask, their architecture's head module estimates a 2-dim gaze direction vector. 

The gaze direction vector is then used to create multiple direction fields, which are empirically generated field-of-view cones represented by a heatmap. These fields are concatenated with the full image, and is fed into a feature pyramid network (FPN) \cite{lin_feature_2017}, followed by a final sigmoid layer to ensure gaze point confidence values fall into standard [0,1] range. This proposed architecture discards the need for a separate scene module, and uses the FPN with sigmoid to perform both the feature extraction and gaze heatmap regression.

\begin{table}[hbp] \centering
\begin{adjustbox}{width=\columnwidth,center}
\begin{tabular}{l|ccc|ccc}
\hline
\multicolumn{1}{c|}{\multirow{2}{*}{Model}} & \multicolumn{3}{c|}{Published}                                           & \multicolumn{3}{c}{Ours}                                                 \\ \cline{2-7} 
\multicolumn{1}{c|}{}                       & AUC $\uparrow$   & Dist. $\downarrow$    &Ang. $\downarrow$ & AUC $\uparrow$   & Dist. $\downarrow$    & Ang. $\downarrow$ \\ \hline
Random                                      & 0.504 & 0.484 & 69.0\degree                                                     & 0.501 & 0.474 & 68.4\degree                                                     \\
Recasens \etal \cite{recasens_where_2015}                           & 0.878 & 0.190 & 24.0\degree                                                     & 0.870 & 0.205 & 28.8\degree                                                     \\
Lian \etal \cite{lian_believe_2019}                         & 0.906 & 0.145 & 17.6\degree                                                     & 0.921 & 0.151 & 18.2\degree                                                     \\
Chong \etal \cite{chong_detecting_2020}                              & 0.921 & 0.137 & n/a                                                        & 0.918 & 0.140 & 17.8\degree                                                     \\ \hline
\end{tabular}
\end{adjustbox}
\caption{Results on GazeFollow Test Set.}
\label{fig: gazefollow_table}
\end{table}

\begin{figure}[t]
\begin{centering}
\includegraphics[width=1.0\columnwidth]{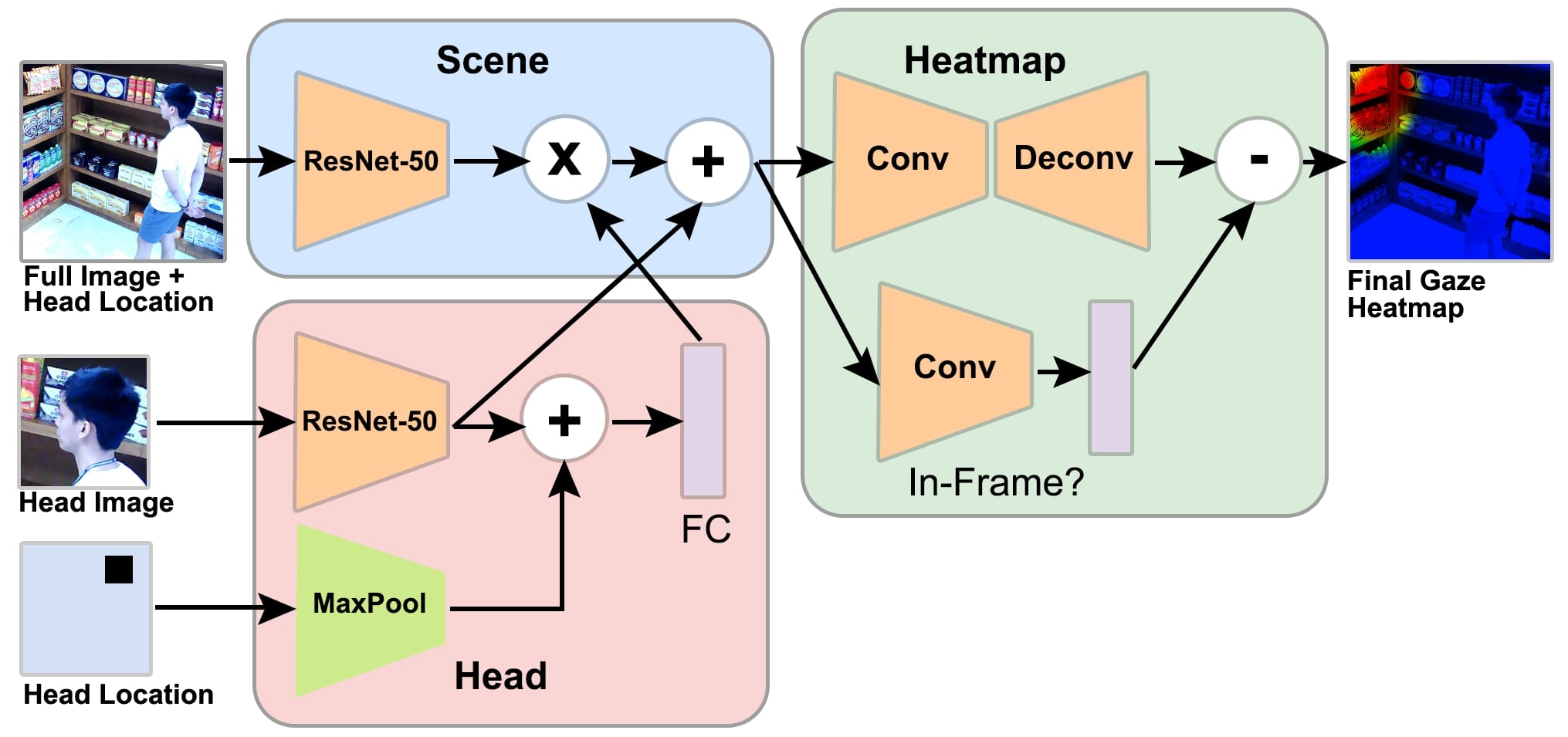}
\par\end{centering}
\vspace{-5px}
\caption{\protect Detecting Attended Visual Targets in Video by Chong \etal \cite{chong_detecting_2020} \label{fig:chong_diag}}
\vspace{-15px}
\end{figure}

\textbf{Detecting Attended Visual Targets In Video.} Chong \etal \cite{chong_detecting_2020} proposed to use both spatial information in static images and temporal information on video to obtain a better gaze heatmap prediction. Their novel architecture introduces a more complex interaction between the head and scene feature maps, as well as convolutional-LSTMs \cite{shi_convolutional_2015} that are able to extract temporal features. Similar to the work of Lian \etal, both the head and scene modules use ResNet-50 to perform feature extraction on the input images. However, their work also introduces additional element-wise connections and operations between the head and scene module, which can be observed in Figure \ref{fig:chong_diag}.


The heatmap module uses two convolutional layers to encode the combined head and scene features. For the purposes of the original authors, a convolutional-LSTM layer comes after the encoding layers for temporal feature extraction. However, for our tasks we only evaluate on static images. Thus, the aforementioned layer is removed. A network composed of three deconvolutional layers and a pointwise convolution upscales the features into a full-sized gaze heatmap. Parallel to this is their novel in-frame branch, which computes a modulating feature map that is subtracted element-wise from the gaze heatmap if it estimates the gaze point to be out of frame. 

\begin{table}[b]
    \centering
\begin{tabular}{l|ccc}
\hline
                Model & AUC $\uparrow$   & Dist. $\downarrow$   & Ang. $\downarrow$ \\ \hline
Random          & 0.497 & 0.454 & 77.0\degree  \\
Recasens \etal \cite{recasens_where_2015} & 0.929 & 0.162 & 33.0\degree \\
Lian \etal \cite{lian_believe_2019}     & \textbf{0.954}  & 0.107 & 19.7\degree \\
Chong \etal \cite{chong_detecting_2020}    & 0.952 & \textbf{0.075} & \textbf{15.1}\degree \\ \hline
\end{tabular}
\vspace{-5px}
\caption{Benchmarking Results on GOO-Synth Test Set.}
\label{fig: synth_table}
\end{table}

\begin{table*}[t] \centering
\begin{tabular}{ll|ccc|ccc}
\hline
\multicolumn{2}{c|}{\multirow{2}{*}{Model}}             & \multicolumn{3}{c|}{No Pretraining}                     & \multicolumn{3}{c}{Pretrained}                         \\ \cline{3-8} 
\multicolumn{2}{c|}{}     & AUC $\uparrow$ & Dist. $\downarrow$ & Ang. $\downarrow$ & AUC $\uparrow$ & Dist. $\downarrow$ & Ang. $\downarrow$ \\ \hline
\multirow{3}{*}{Recasens} & \multicolumn{1}{c|}{0-shot} & 0.543          & 0.359              & 78.2              & 0.706          & 0.313              & 74.0              \\
                          & 1-shot                      & 0.746          & 0.263              & 49.7              & 0.872          & 0.196              & \textbf{38.5}     \\
                          & 5-shot                      & 0.850          & 0.220              & 44.4              & \textbf{0.903} & \textbf{0.195}     & 39.8              \\ \hline
\multirow{3}{*}{Lian}     & 0-shot                      & 0.502          & 0.420              & 69.2              & 0.773          & 0.275              & 49.6              \\
                          & 1-shot                      & 0.723          & 0.688              & 71.2              & 0.866          & 0.178              & 34.4              \\
                          & 5-shot                      & 0.840          & 0.321              & 43.5              & \textbf{0.890} & \textbf{0.168}     & \textbf{32.6}     \\ \hline
\multirow{3}{*}{Chong}    & 0-shot                      & 0.670          & 0.334              & 66.6              & 0.710          & 0.255              & 47.9              \\
                          & 1-shot                      & 0.723          & 0.301              & 63.2              & 0.839          & 0.188              & 36.0              \\
                          & 5-shot                      & 0.796          & 0.252              & 51.4              & \textbf{0.889} & \textbf{0.150}     & \textbf{29.1}     \\ \hline
\end{tabular}
\caption{Performance on GOO-Real Test set. Models that receive pretraining on GOO-Synth before being few-shot trained on GOO-Real are compared to their performance when GOO-Synth pretraining is skipped.}

\label{fig: real_table}
\vspace{-8px}
\end{table*}

\section{Experiments}

We evaluate the performance of the methods discussed in Section \ref{methodology} on the tasks of gaze following and domain adaptation. Several baselines \cite{recasens_where_2015, lian_believe_2019, chong_detecting_2020} are initially benchmarked on the GazeFollow \cite{recasens_where_2015} dataset to check the accuracy of our replication when compared to the results achieved in their respective publications. We then present the benchmarks of these methods on the GOO-Synth and GOO-Real datasets. We leave experimentation with new architectures, loss functions, and metrics for gaze object prediction to future work. 

\subsection{Implementation Details}

All baseline methods are implemented in a unified, modular codebase based on the PyTorch framework. Training and evaluation of networks are performed on a single machine using a GeForce GTX 1080Ti. All necessary pretraining and initialization methods are lifted from each method's respective publications to recreate results as accurately as possible. In the absence of disclosed training hyper-parameters such as in the case of \cite{recasens_where_2015}, training is empirically tuned to obtain values nearest to the original implementation. We also made the codebase available in the interest of reproducibility and future work.

\subsection{Evaluation}

The standard metrics for evaluating gaze following are used not only for the GazeFollow dataset, but also for the GOO dataset. We consider the standard metrics to be as follows: Area Under the ROC Curve (\textbf{AUC}) is implemented as described in prior work \cite{judd_learning_2009}, where the prediction and ground truth heatmap are downscaled and used as confidence values to produce an ROC curve. $L_{2}$ distance (\textbf{Dist.}) is the euclidean distance between the predicted and ground truth gaze point when the image dimensions is normalized to 1$\times$1. Angular error (\textbf{Ang.}) is the angular difference between the gaze vectors when connecting the head point to the predicted and ground truth gaze points.


Given the synthetic and real partitioning of the GOO dataset, experiments on domain adaptation through simple transfer learning are conducted. Models which have been trained until convergence on GOO-Synth are subjected to 0-shot, 1-shot, and 5-shot training on GOO-Real before being evaluated on its test set. Models which have not been given GOO-Synth pretraining are also trained with the same hyper-parameters, and quantitative comparisons between the two setups are made using the previously discussed metrics.

\begin{figure*}[t]
\begin{centering}
\includegraphics[width=\linewidth]{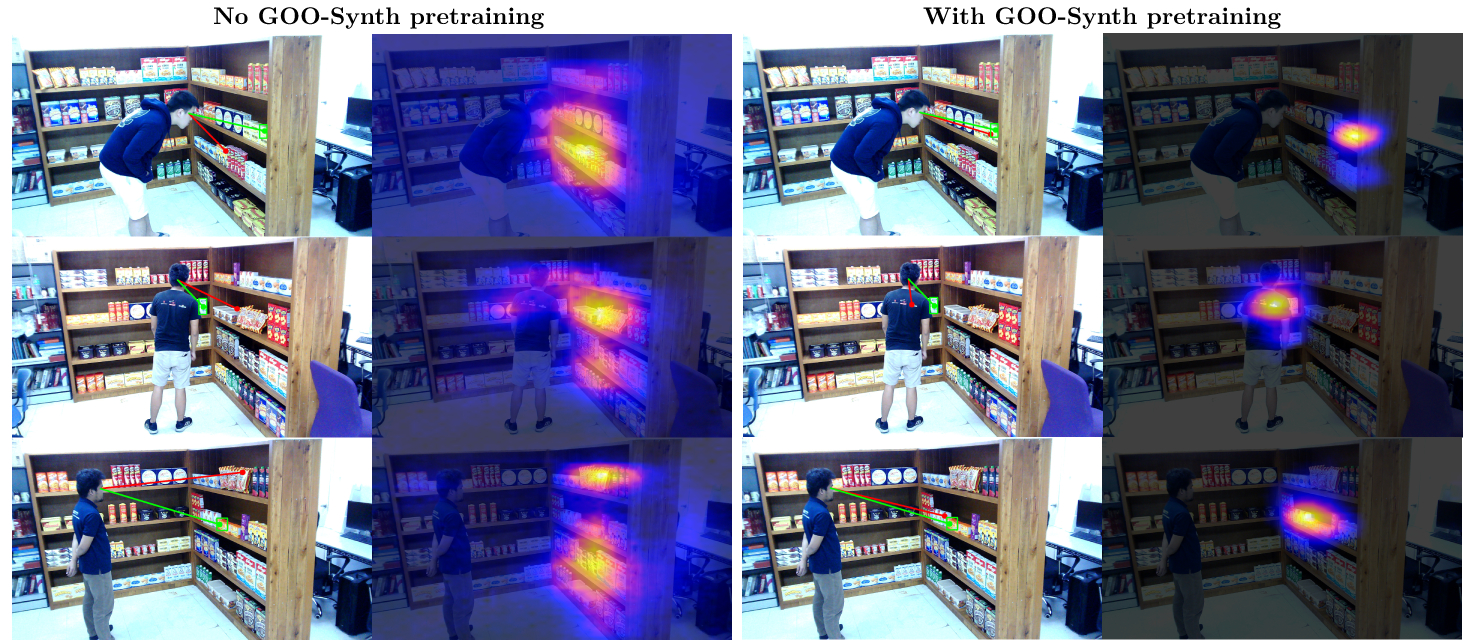}
\par\end{centering}
\caption{\protect Sample predicted points and heatmaps using Chong \etal's gaze network. Green line represents the ground truth gaze vector and gaze object bounding box, while the red line is the model prediction. When evaluated on GOO-Real, models that have been pretrained on GOO-Synth produces more precise heatmaps than models only pretrained with GazeFollow. \label{fig:chong_output}}
\end{figure*}

\subsection{Results \& Analysis}

\textbf{GazeFollow.}
Shown in Table \ref{fig: gazefollow_table} is our re-implementation of the discussed algorithms and their benchmarks in comparison with the results published in their respective papers. Our version of Recasens \etal has the greatest discrepancy between the authors' results and ours, which we attribute to lack of training details provided in the paper, in addition to their model being implemented in a different framework. The performance of the works of Lian \etal and Chong \etal achieves much more accurate values due to the respective authors making their code available online. 

\textbf{GOO-Synth.} We present the results on the GOO-Synth dataset for both the gaze following and gaze object prediction task,  shown in Table \ref{fig: synth_table}. By comparing the benchmarks achieved on GOO-Synth to the results on GazeFollow, some analysis can be drawn regarding the differences in context between the two datasets. On the task of gaze following, baselines achieve higher performing values on AUC and L2 distance. This is hypothesized to be because of the singular context of retail for the GOO scenes as opposed to the varying scene context of GazeFollow data, making point estimates and heatmaps easier to learn for the models. On angular error however, baselines perform worse on the GOO-Synth dataset. We determine this to be the effect of images in GOO where the human head is facing opposite the camera but towards the shelves, making it hard for models to make use of head features to estimate direction. In summary, the scene module of the baselines perform better on the GOO dataset where only the retail scenario exists, while the head module performs slightly worse due to cases where the head is facing away from the camera. 


\textbf{GOO-Real.} Results for baseline evaluation on GOO-Real can be observed in Table \ref{fig: real_table}. The values consistently show how models trained on the GOO-Synth dataset before being trained on GOO-Real achieve higher performance on all metrics compared to models without. Performance shown in 0-shot by pretrained models indicate better initialization of model weights across all the baselines; 1-shot evaluation shows that these models achieve competitive performance with less training iterations; and lastly, 5-shot training results imply that GOO-Synth pretrained models are able to adapt the learned synthetic features to obtain higher performance approaching convergence.  

\textbf{Qualitative.} Sample gaze point and heatmap predictions using Chong \etal's gaze network are shown in figure \ref{fig:chong_output}. After 5-shot training and evaluation on GOO-Real, models with pretraining on GOO-Synth achieve higher quality heatmaps and more precise point predictions. Models with no GOO-Synth pretraining seems to be unable to confidently classify background pixels, producing the blue tint on the heatmap outputs in column 1. The sample in row 2 implies that the GOO-Synth pretrained model is more robust to subjects with their back and head completely turned away from the camera. The model initialized only with GazeFollow tends to produce heatmaps with multiple hotspots, which was alleviated by the synthetic pretraining as reflected in row 3.

\section{Conclusion}

In this paper, we present Gaze On Objects (GOO), a dataset for gaze object prediction set in a retail environment, consisting of 192,000 images from a simulated environment (GOO-Synth) and 9,552 images from a real-world setup (GOO-Real). We introduce the task of gaze object prediction, which would hopefully inspire novel architectures and training methods for gaze systems to infer the class and boundaries of the specific object being looked at. We provide thorough baseline experiments for benchmarking existing gaze following methods on our dataset. Our work also provides a comprehensive evaluation of networks on GOO-Real given whether they were pretrained on GOO-Synth or not, in the interest of domain adaptation.  

\section{Future Work}

The benchmarks shown in this paper focused only on existing metrics on gaze estimation. However, to fully complete the gaze on objects task, it is necessary to also formulate new metrics to measure the performance of predicting the gaze object. This includes measuring the correctness of both the bounding box and the class of the object. This was excluded since the current works on gaze estimation do not predict bounding boxes nor classes. We hope to include this in future work on gaze object prediction.

\section{Acknowledgement}
This work was funded by Samsung R\&D Institute Philippines. Special thanks to the people of Computer Networks Laboratory: Roel Ocampo, Vladimir Zurbano, Lope Beltran II, and John Robert Mendoza, who worked tirelessly during the pandemic to ensure that our network and servers are continuously running.  

{\small
\bibliographystyle{ieee_fullname}
\nocite{*}
\bibliography{GOO_dataset}
}


\end{document}